\documentclass[letterpaper, 10 pt, conference]{ieeeconf}  

\IEEEoverridecommandlockouts                              

\usepackage{atbegshi,picture}
\usepackage{float}
\usepackage{graphics} 
\usepackage{epsfig} 
\usepackage{mathptmx} 
\usepackage{times} 
\usepackage{amsmath} 
\usepackage{amssymb}  
\usepackage{import}
\usepackage{colortbl}
\usepackage{wrapfig,lipsum}
\usepackage{graphicx}
\usepackage{booktabs}
\usepackage{multirow}
\usepackage{placeins}
\usepackage{gensymb}
\usepackage{xcolor}
\usepackage{soul}
\usepackage{textcomp}
\usepackage[ruled,linesnumbered, noend]{algorithm2e}
\usepackage{amsmath,amssymb}
\DeclareMathOperator*{\argminB}{argmin}   
\usepackage{tikz}
\usepackage{yfonts}
\usepackage{graphicx}
\usepackage{cite}
\usepackage{hyperref}
{\end{list}}

\definecolor{Gray}{gray}{0.85}
\newcolumntype{a}{>{\columncolor{Gray}}c}

\title{\LARGE \bf $se(3)$-TrackNet: Data-driven 6D Pose Tracking\\
by Calibrating Image Residuals in Synthetic Domains}

\author{Bowen Wen, Chaitanya Mitash, Baozhang Ren, Kostas E. Bekris
\thanks{Work by the authors has been supported by NSF awards 1723869 and 1734492. The authors are with the Computer Science Department of Rutgers University in Piscataway, New Jersey, 08854, USA. Email: {\tt\small \{bw344, cm1074, kb572\}@rutgers.edu}}%
}

\begin{document}

\AtBeginShipoutNext{\AtBeginShipoutUpperLeft{%
  \put(\dimexpr\paperwidth-0.5cm\relax,-0.5cm){\makebox[0pt][r]{\framebox{Accepted in International Conference on Intelligent Robots and Systems (IROS) 2020}}}%
}}

\maketitle
\thispagestyle{empty}
\pagestyle{empty}

\begin{abstract}
Tracking the 6D pose of objects in video sequences is important for  robot manipulation. This  task, however, introduces multiple challenges: (i) robot manipulation involves significant occlusions; (ii) data and annotations are troublesome and difficult to collect for 6D poses, which complicates machine learning solutions, and (iii) incremental error drift often accumulates in long term tracking to necessitate re-initialization of the object's pose. This work proposes a data-driven optimization approach for long-term, 6D pose tracking. It aims to identify the optimal relative pose given the current RGB-D observation and a synthetic image conditioned on the previous best estimate and the object's model. The key contribution in this context is a novel neural network architecture, which appropriately disentangles the feature encoding to help reduce domain shift, and an effective 3D orientation representation via Lie Algebra. Consequently, even when the network is trained only with synthetic data can work effectively over real images. Comprehensive experiments over benchmarks - existing ones as well as a new dataset with significant occlusions related to object manipulation - show that the proposed approach achieves consistently robust estimates and outperforms alternatives, even though they have been trained with real images. The approach is also the most computationally efficient among the alternatives and achieves a tracking frequency of 90.9Hz. \footnote{Code, data and supplementary video for this project are available at \href{https://github.com/wenbowen123/iros20-6d-pose-tracking}{https://github.com/wenbowen123/iros20-6d-pose-tracking}}
\end{abstract}

\section{INTRODUCTION}


Robotic tasks, such as object manipulation, often require to track the pose of an object.  Pose estimation from a single snapshot can initiate a manipulation pipeline and has been
studied extensively \cite{xiang2017posecnn, deng2019poserbpf, mitash2019scene,deng2019self, 
issac2016depth, sundermeyer2018implicit,wang2019densefusion,li2019cdpn,he2020pvn3d}. Purposeful manipulation, however, such as placement and especially within-hand reorientation \cite{kimmel2019belief}, critically depends on online tracking \cite{mitash2020task}.  Some pose estimation approaches are relatively fast and can re-estimate pose from scratch for every frame \cite{tremblay2018deep, wang2019densefusion, sundermeyer2018implicit,wen2020robust}. This can be redundant, however, and often leads to less coherent estimations over consecutive frames, which negatively impact manipulation. 

\begin{figure}[t]
  \centering
  \includegraphics[width=0.48\textwidth,height=0.2\textwidth]{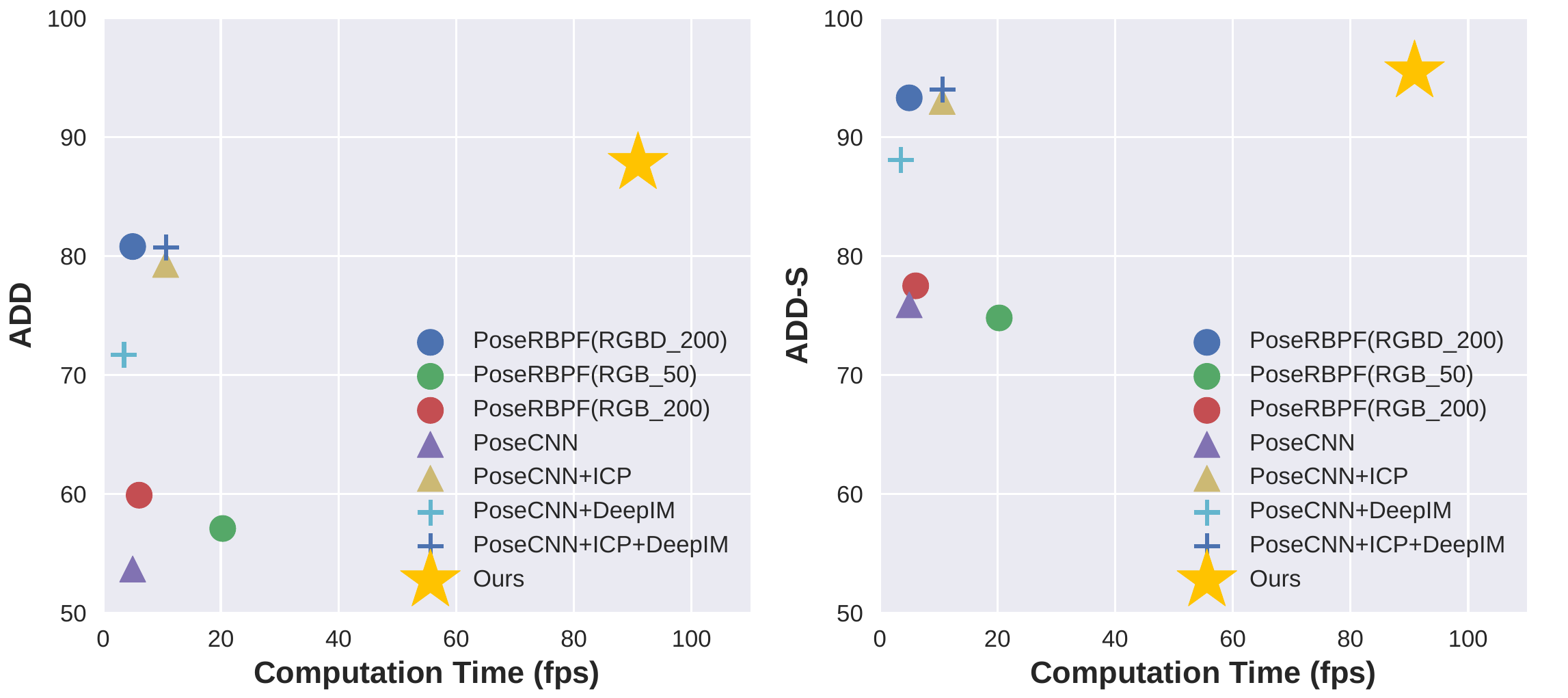}
  \includegraphics[width=0.48\textwidth]{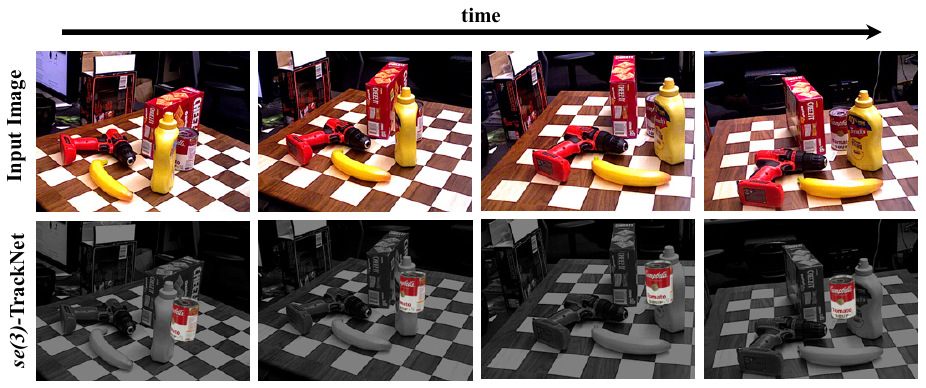}
  \vspace{-.3in}
\caption{\textbf{Top:} Performance w.r.t. computation time evaluated on
the YCB-Video dataset according to the area under the curve (AUC)
metric for the ADD and ADD-S objectives \cite{xiang2017posecnn}. The
proposed approach is able to perform more accurate tracking while
being significantly faster than alternatives. \textbf{Bottom:} Pose
predicted by the $se(3)$-TrackNet without any re-initialization, which is
able to recover from complete occlusion.}
\label{fig:acc_speed}
  \vspace{-.3in}
\end{figure}

Temporal tracking of object poses over sequences of images can greatly
improve speed while maintaining or even improving pose
quality \cite{Wthrich2013ProbabilisticOT, schmidt2014dart}. Nevertheless, many
traditional methods, which depend on hand-crafted likelihood functions
and features, require extensive hyper-parameter tuning when adapt to 
novel object categories or environments. On the other hand,
data-driven techniques \cite{li2018deepim,garon2017deep} require 
real-world training data, which are difficult to acquire and
label in the context of 6D poses.

This work proposes a data-driven optimization strategy to keep long-term track of an object's pose robustly, as shown in Fig. \ref{fig:acc_speed}. The
contributions are the following:

\noindent 1. A novel deep neural network that learns to predict the relative pose between the current observation and the synthetic model rendering at
the previous prediction. A smart feature-encoding disentanglement technique enables more efficient sim-to-real transfer.

\noindent 2. A Lie Algebra representation of 3D orientations, which allows effective learning of the residual pose transforms given a proper loss function. 

\noindent 3. A training pipeline over synthetic data that employs 
domain randomization \cite{tobin2017domain} in the context of pose tracking. Automatic training data generation significantly reduces the manual effort in collecting and labeling videos for tracking 6D poses.

\noindent 4. A novel benchmarking dataset for 6D pose tracking in the context of multiple different robotics manipulation tasks. It was collected with various robotic end-effectors and YCB objects, where 6D object pose annotations are provided in every frame of the video.

Experiments indicate that the proposed network achieves state-of-art
results on the YCB-Video benchmark without re-initialization in contrast to prior work \cite{li2018deepim,deng2019poserbpf}. It is also significantly
faster at 90.9Hz. This allows use in real-time scenarios such as robot manipulation.


\section{RELATED WORK}
\noindent \textbf{Data-driven 6D Pose Estimation:} Learning-based techniques have shown promise in directly regressing the 6D object pose from image data \cite{xiang2017posecnn,
  wang2019densefusion}. Nevertheless, given the complexity of the 6D
challenge, a large amount of pose annotated training data is required
to achieve satisfactory results in practice. This is often more
challenging than labeling for object classification or detection. Some
data-driven techniques combine deep learning with traditional
approaches like PnP \cite{tremblay2018deep}. Although
more data-efficient, this could be problematic under severe
occlusions, and often requires precisely calibrated camera parameters
for the PnP step. Given that a pose is re-estimated in every frame,
estimation techniques often trade-off speed for accuracy
\cite{xiang2017posecnn}. This might not be
desirable for manipulation. In contrast, the current work exploits
temporal information to achieve higher accuracy and faster response than 
state-of-art single-image pose estimation methods while using only synthetic data for training.

\noindent \textbf{6D Pose Tracking:} For setups where CAD object
models are available, the approaches can be generally categorized to probabilistic and optimization-based.
 \textit{Probabilistic Tracking:} An efficient particle filtering
framework \cite{choi2013rgb} harnesses the computational power of GPU,
where likelihood is computed based on color, distance and
normals. Nevertheless, the hand-designed likelihood functions can hardly generalize to different lighting conditions or scenes with challenging
clutter. An alternative \cite{Wthrich2013ProbabilisticOT} explicitly
models occlusions and shows success in terms of robustness but the
pose accuracy is not precise enough for certain manipulation tasks. To
ameliorate this problem, follow up work \cite{issac2016depth} applied
Gaussian Filtering and achieves promising pose accuracy but introduces
more frequent tracking loss. Recent work \cite{deng2019poserbpf}
proposed a Rao-Blackwellized particle filter, which decouples the
translational and rotational uncertainty, achieving state-of-art 6D
pose tracking performance on the YCB Video benchmark. In the case of
severe occlusions, however, re-initialization of the pose estimation
is required. \textit{Optimization-based Tracking:} A number of methods proposed objective functions, which capture the
discrepancy between the current observation and the previous state.
They compute relative transformations based on the minima of the
residual function \cite{schmidt2014dart, pauwels2015simtrack,
  joseph2015versatile, zhong2019robust, tjaden2018region}. In
particular, methods combine the optical/AR flow and point-to-plane distance to solve
tracking in a least-squares sense \cite{pauwels2015simtrack, schmidt2014dart}. SIFT features and optical flow noise, however, limit performance and often require extensive
hyper-parameter tuning to adapt to new scenarios. The most related effort to the current paper leverages the FlowNetSimple network
\cite{fischer2015flownet} to refine pose outputs of any 6D object pose
detection approach and can also be extended to tracking \cite{li2018deepim}. It requires,
however, occasional re-initialization and has to be trained at least
partially with real data.

\noindent \textbf{Simulation to Reality:} Training on synthetically generated datasets allows faster, more scalable, and lower-cost data collection \cite{tobin2017domain}. The
discrepancy between the synthetic and real
data, however, often results in significant performance drop. Domain adaptation
techniques like Gradient-reversal \cite{ganin2014unsupervised} and
utilizing Generative Adversarial Networks (GANs) for input space
domain alignment \cite{shrivastava2017learning} help bridge this
gap. These methods, however, often assume the source domain already
resembles the target domain to certain extent, which could not be
trivially satisfied in practice. Advancements in computer graphics, \cite{yosinski2014transferable,movshovitz2016useful}
have shown the
benefits of performing photo-realistic renderings on tasks. Achieving such photo-realism, however, often introduces another source of
human involvement and expert domain knowledge.  Domain randomization efforts impose that the rendering settings of the simulator are randomized and certain transferability to real world has been demonstrated \cite{tobin2017domain}. This work inherits the idea of domain randomization but also pursues physical plausibility. Together with feature encoding disentanglement, the proposed network can be trained exclusively over synthetic data and is shown to generalize to the real world.

\begin{figure}[h]
  \centering
  \includegraphics[width=\linewidth]{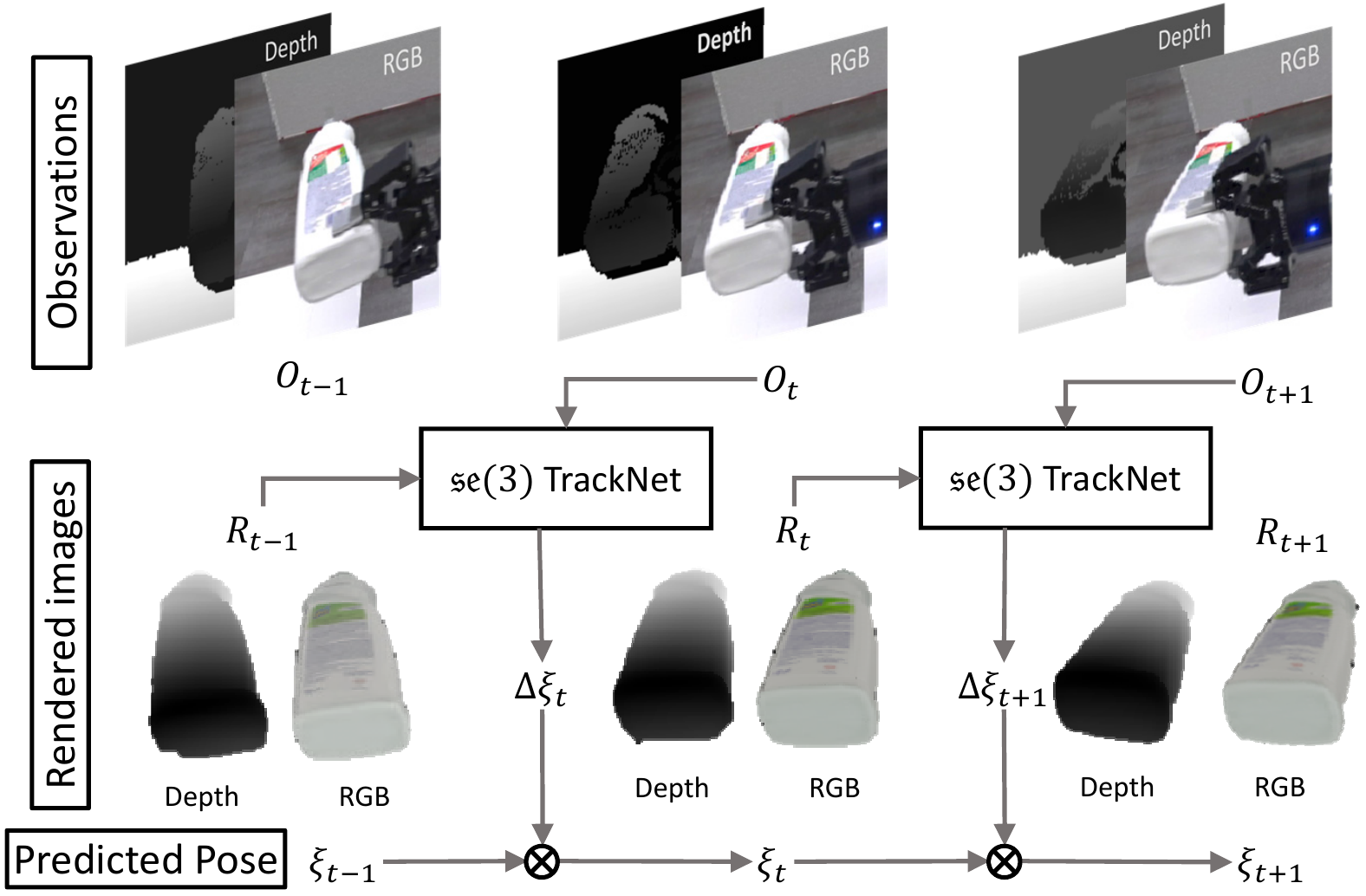}
  \vspace{-.3in}
  \caption{Overview: At any given time $t$ the current observation $O_t$ and the rendering $R_{t-1}$ of the object model based on the previously computed pose $\xi_{t-1}$ are passed to the $se(3)$-TrackNet. The network computes the relative pose $\Delta\xi_{t}$, which is then propagated forward to compute $\xi_t$.}
  \label{fig:overview}
\end{figure}



\section{APPROACH}
The objective of this work is to compute the 6D pose of an object $T_t \in SE(3)$ at any time $t > 0$, given as input:
\begin{itemize}
    \item a 3D {\tt CAD} model of the object,
    \item its initial pose, $T_0 \in SE(3) $, computed by any single-image based 6D pose estimation technique, and a
    \item sequence of RGB-D images $O_{\tau }, \tau \in \{0,1,...,t-1\}$ from previous time stamps and the current observation $O_t$.
\end{itemize}
This work proposes a data-driven optimization technique to track the object pose over RGB-D image sequences. The cost function for the optimization is encoded and learned by a novel neural network architecture, trained with only synthetically generated data. In every time step, the proposed approach, computes a residual over the pose computed for the object model in the previous frame as indicated in Fig. 2. The details of this formulation, the neural network architecture and the data generation pipeline are provided below.

\begin{figure*}[t]
  \centering
  \includegraphics[width=0.95\textwidth]{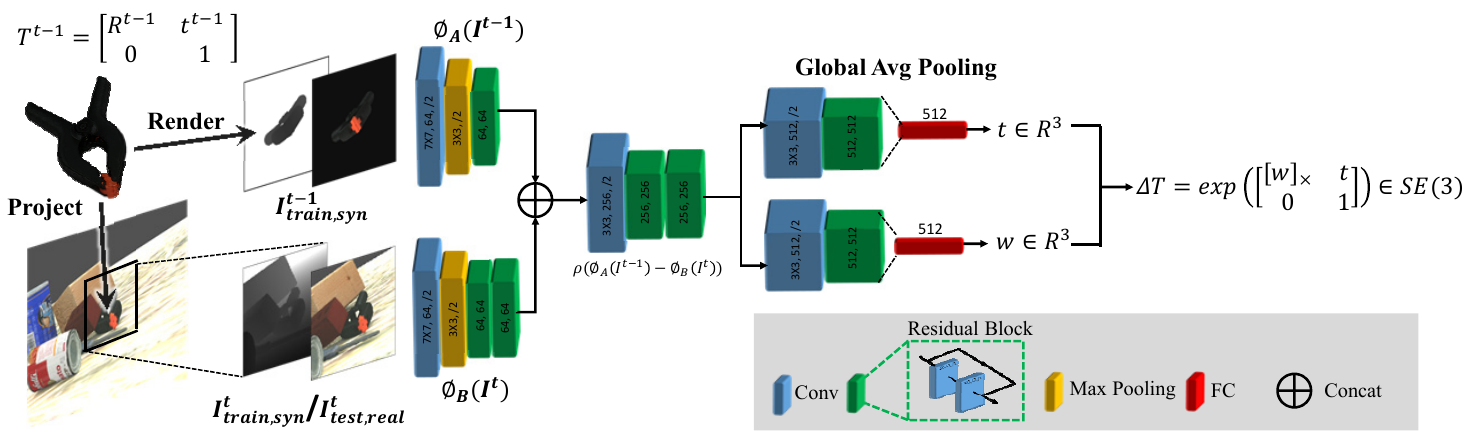}
  \vspace{-.15in}
  \caption{Proposed $se(3)$-TrackNet architecture: It takes as input RGB-D images corresponding to the current observation and a rendering of the object model at the previous timestamp, into two separate feature encoders $\phi_B$ and $\phi_A$ respectively. Both inputs are synthetic during training while at test time, the input to $\phi_B$ is a real image. The encoders' outputs are concatenated and used to predict the relative pose between the two images, with decoupled translation and rotational. \vspace{-0.2in}}
  \label{fig:pipeline}
\end{figure*}


\subsection{Tracking on $SE(3)$ Manifolds with Residuals}
\label{sec:se3}

Optimization in this domain operates over cost functions defined for the object poses $\bar{\xi}, \xi$ that measure the discrepancy $\epsilon$ between the features extracted from the images:
$$\varepsilon = \rho(\phi_{I_1} (\bar{\xi}) - \phi _{I_2}(\xi)),$$ where $\rho$ is a predefined robust loss function, and $\phi(\cdot)$ can be direct pixel intensity values, such as in \cite{engel2017direct}, point-to-point discrepancy or its variations \cite{besl1992method}, pre-designed features \cite{hebert2012combined} or the combinations from any of the above \cite{pauwels2015simtrack}. 

Given the current observation $O_t$, and the pose computed in the previous timestamp $\xi_{t-1}$, the goal of this work is to find a relative transformation $\Delta \xi$ that takes the object from $\xi_{t-1}$ to the pose captured by the current observation. This can be formulated as an optimization problem. Let $R$ denote the image corresponding to the object model's rendering at the given pose. Then, the optimal relative transform is:
\vspace{-0.05in}
$$
\Delta \xi^* = \argminB_{\Delta \xi} \{\rho(\phi_{O_{t}}(\xi_t) - \phi_R(\xi_{t-1} \boxplus \Delta \xi))\}
\vspace{-0.05in}
$$
A general approach to solve this is to perform Taylor expansion around $\xi$, which reformulates the equation as $\phi_{R} (\xi_{t-1} \boxplus \Delta \xi) = \phi_R(\xi_{t-1}) + J(\xi_{t-1}) \Delta \xi$, where $J$ is the Jacobian matrix of $\phi_R$ w.r.t. $\xi$. Now $\xi$ is locally parametrized in its tangent space, specifically $\xi=(t,w)^T \in se(3)$, such that its exponential mapping lies in the Lie Group
$\Delta T=exp(\Delta \xi)=\begin{bmatrix}
R &t \\ 
0 & 1
\end{bmatrix} \in SE(3)
$, where $R=I_{3\times3}+\frac{[w]_{\times}}{|w|}sin(|w|)+\frac{[w]_{\times}^2}{|w|^2}(1-cos(|w|))$ and $[w]_{\times}$ is the skew-symmetric matrix.

In the case of $L_2$ loss function without loss of generality:
\vspace{-0.1in}
$$
\Delta \xi = (J^T J)^{-1} J^T  || (\phi_{O_t} (\xi_{t}) - \phi_{R} (\xi_{t-1} \boxplus \Delta \xi)) ||.
\vspace{-0.05in}
$$
Solving the expression by explicitly deriving the Jacobian matrix and iteratively updating often requires a formalized cost function or the features extracted from the observations to be differentiable w.r.t. $\xi$, and appropriate choice of a robust cost function and hand-crafted features. Another problem arises when different modalities are involved. In such cases, another hyper-parameter controlling the importance of each modality (e.g., RGB-D) has to be introduced \cite{pauwels2015simtrack} and could become non-trivial to tune for all different scenarios. 

Instead, this work proposes a novel neural network architecture that implicitly learns to calibrate the residual between the features extracted from the current observation and the rendered image conditioned on previous pose estimate to resolve the relative transform in the tangent space $\Delta \xi \in se(3)$. 

\subsection{Neural Network Design}
The proposed neural network is shown in Fig. \ref{fig:pipeline}. The network takes as input a pair of images, $I^{t-1}$: rendered from the previous pose estimation, and $I^{t}$: the current observation. The images are 4-channel RGB-D data. Depth is often available in robotics. Nevertheless, it complicates learning due to the additional domain-gap between synthetic and real depth images. In addition, not all neural network architectures are well suited to encode RGBD features, such as the FlowNetSimple architecture \cite{fischer2015flownet} used in \cite{li2018deepim}.

During training, both inputs are synthetically generated images $\phi(I_{syn,train}^{t-1};I_{syn,train}^{t})$, while for testing the current timestamp input comes from a real sensor, $\phi(I_{syn,test}^{t-1};I_{real,test}^{t})$. The $se(3)$-TrackNet uses two separate input branches for $I^{t-1}$ and $I^{t}$. The weights of the feature encoders are not shared so as to disentangle feature encoding. This is different from related work \cite{li2018deepim}, where the two images are concatenated into a single input. A shared feature extractor worked in the previous work when both real and synthetic data are available during training. The property of the latent space $\phi(I_{syn,train}^{t-1};I_{real,train}^{t})$ could still be partly preserved when tested on real world test scenarios $\phi(I_{syn,test}^{t-1};I_{real,test}^{t})$. This representation, however, does not generalize to training exclusively on synthetic data. 

The latent space features trained on purely synthetic data are denoted as $\phi_A(I_{syn,train}^{t-1})$ and $\phi_B(I_{syn,train}^{t})$. When tested on real world data, the latent space features are $\phi_A(I_{syn,test}^{t-1})$ and $\phi_B(I_{real,test}^{t})$. By this feature encoding disentanglement, domain gap reduces to be between $\phi_B(I_{syn,train}^{t})$ and $\phi_B(I_{real,test}^{t})$, while $\phi_A(I_{syn,train}^{t-1})$ and $\phi_A(I_{syn,test}^{t-1})$ can be effortlessly aligned between the training and test phase without the need for tackling the domain gap problem. 

\begin{figure}[t!]
  \centering
  \includegraphics[width=0.45\textwidth]{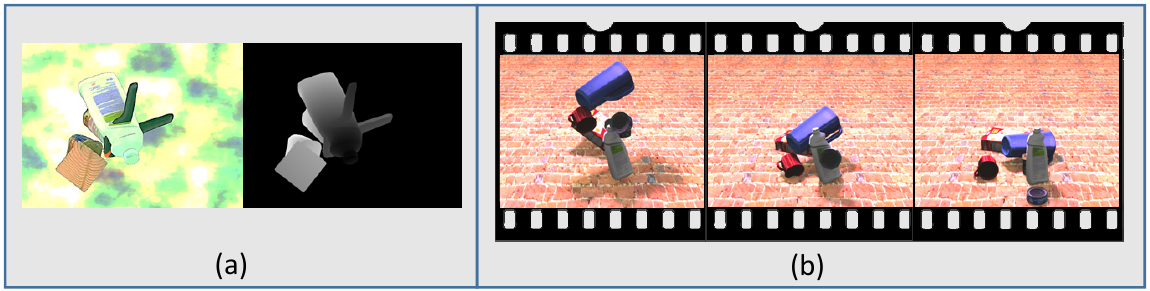}
  \vspace{-.1in}
  \caption{Comparison of Domain Randomization (DR) against  Physically Plausible Domain Randomization (PPDR). (a) DR directly renders using a sampled object pose. Notice the penetration between objects, which can introduce undesired bias to depth data. (b) In PPDR, a randomly sampled pose serves as initialization for physics simulation. Rendering is performed over the stable object pose. The domain invariant, penetration-free property can then help to effectively align the synthetic and real domains. \vspace{-0.2in}}
  \label{fig:PPDR}
\end{figure}

A relative transform can be predicted by the network via end-to-end training. The transformation is represented by Lie algebra as $\Delta \xi = (t,w)^T  \in se(3)$, where the prediction of $w$ and $t$ are disentangled into separate branches and trained by $L_2$ loss: 
\vspace{-0.15in}
$$
L=\lambda_1||w-\bar{w}||_2 + \lambda_2||t-\bar{t}||_2
\vspace{-0.04in}$$
\noindent where $\lambda_1$ and $\lambda_2$ has been simply set to 1 in experiments. Given $\Delta \xi$, the current pose estimate is computed as  $T^{t}=exp(\Delta \xi) \cdot T^{t-1}$, as described in Sec. \ref{sec:se3}. 

\label{sec:syn_data_gen}
\subsection{Synthetic Data Generation via PPDR}

The purpose of domain randomization is to provide enough simulated variability at training time, such that at test time, the model is able to generalize to real-world data \cite{tobin2017domain}. Prior work implements the idea of domain randomization by randomly changing the number of objects, poses, textures, lighting, etc, where object poses are usually directly sampled from some predetermined distribution \cite{tobin2017domain,xiang2017posecnn,tremblay2018deep}. Although it is non-trivial to align some complex physical properties between the simulator and real world, such as lighting and camera properties, certain physical properties, for instance gravity and collision can be effortlessly preserved \cite{tremblay2018falling,mitash2017self}. This makes domain invariant features more tractable to be captured by the neural network. It is especially important in the current framework when depth modality is additionally employed and unrealistic object penetration which never occurs in real world introduces undesired bias to depth data distribution during learning.

This work therefore leverages the complementary attributes of domain randomization and physically-consistent simulation for the synthetic data generation process. The goal is to combine the two ideas such that the synthetic training data holds a diverse distribution for the network to generalize to the target domain with different environments, while being more data-efficient. We refer to this idea as PPDR (Physically Plausible Domain Randomization). More concretely, object poses are initialized randomly where collision between objects or distractors could occur, which is then followed by a number of physics simulation steps so that objects are separated or fall onto the table without collision. Other complex or intractable physical properties such as lighting, number of objects, distractor textures are randomized. A comparison between Domain Randomization against PPDR is illustrated in Fig. \ref{fig:PPDR}. 

Once the synthetic image is generated for the entire scene, paired data $I^{t-1}_{syn,train}$ and $I^{t}_{syn,train}$ are extracted and utilized as the input to the network. $I^{t}_{syn,train}$ is obtained by cropping the image given the target object's dimension and zoomed into a fixed resolution $176 \times 176$ before feeding into the network - similar to prior work \cite{li2018deepim}. $I^{t-1}_{syn,train}$ is obtained by randomly sampling a perturbated pose $T_{t-1}^{t}$ where its translation's direction is uniformly sampled and its norm follows a Gaussian distribution $|t| \sim |N(0,\sigma_t)|$ . The rotation is locally parameterized in the tangent space $w \in so(3)$ as discussed above and the direction of $w$ is also uniformly sampled, while its norm is sampled from a Gaussian distribution $|w| \sim |N(0,\sigma_w)|, w \in R^3$, similar to $t$.

The next step within this context is to {\it bridge the domain gap of depth data via bidirectional alignment}. Similar to the case of RGB, sim-to-real gap also arise in terms of the depth data, especially for those captured by a commercial-level depth sensor. However, there has been less evidence about how the depth domain gap could be resolved in a general way, especially that it could be partly dependent on the specific depth sensor. In this work, a bidirectional alignment is performed between the synthetic depth data during training time and real depth data during test time. Specifically, during training time, two additional data augmentation steps are applied to the synthetic depth data $D^{t}_{syn,train}$ at branch B. First, random Gaussian noise is added to the pixels with a valid depth value, which is then followed by a depth-missing procedure by randomly changing part of pixels with valid depth into invalid so as to resemble a real corrupted depth image captured by commercial-level depth sensors. In contrast, during test time, a bilateral filtering is carried out on the real depth image so as to smooth sensor noise and fill holes to be aligned with the synthetic domain.

\section{EXPERIMENTS}
This  section  evaluates  the  proposed  approach  and  compares  against  state-of-the-art 6D pose tracking methods as well as single-image pose estimation methods on a public benchmark. It also introduces a new benchmark developed as part of this work, which corresponds to robot manipulation scenarios. Extensive experiments are performed over diverse object categories and various robotics manipulation scenarios (moving camera or moving objects). Both quantitative and qualitative results demonstrate the advantages of the proposed method in terms of accuracy and speed, while using only synthetic training data. Except for training, all  experiments  are  conducted  on  a  standard  desktop with  Intel  Xeon(R) E5-1660 v3@3.00GHz  processor. Neural network training and inference are performed on a  NVIDIA RTX 2080 Ti GPU and  NVIDIA Tesla K40c GPU respectively.

The synthetic data generation pipeline is implemented in \textit{Blender}\footnote{https://www.blender.org/}. To render images, the camera's pose is randomly sampled from a sphere of radius between 0.6 to 1.3 m, followed by an additional rotation along camera z-axis sampled between 0 to $360^{\circ}$. The number of external lighting sources (lamps) is sampled within 0 to 2 with varying poses. The strength and color of the environment and the lights are randomized. Object poses are randomly initialized, followed by physics simulation, which is terminated after 50 steps to ensure objects have been separated without collisions or fallen onto the table. For {\it YCB-Video}, table textures are randomly selected from \cite{galerne2010random}. For {\it YCBInEOAT}, tables are removed and background is replaced by images captured in the same location, where the data were collected. For each pair $I^{t-1}_{syn,train}$ and $I^{t}_{syn,train}$, their relative transformation $T_{t-1}^{t}$ is sampled following a Gaussian distribution as described in Sec. \ref{sec:syn_data_gen}, where $\sigma_t$ and $\sigma_w$ are empirically set to 2 cm and $0.262\ rad \ (=15^{\circ} )$ respectively. 200k data points (image pairs) are used for training the training set. The network is trained with Adam optimizer for 300 epochs with a batch size of 200. Learning rate starts from 0.001 and is scaled by 0.1 at epochs 100 and 200. Input RGB-D images are resized to $176 \times 176$ before sending to the network. Data augmentations including random HSV shift, Gaussian noise, Gaussian blur are added only to $I^{t}_{syn,train}$. Additional depth-missing corruption augmentation is applied to $D^{t}_{syn,train}$ as described in Sec. \ref{sec:syn_data_gen} by a missing percentage between 0 to 0.4. For both training and inference, rendering of $I^{t-1}$ is implemented in C++ OpenGL.

\subsection{Datasets}
\noindent \textbf{YCB-Video Dataset} This dataset \cite{xiang2017posecnn} captures 92 RGB-D video sequences over 21 YCB Objects \cite{calli2015ycb} arranged on table-tops. Objects' groundtruth 6D poses are annotated in every frame. The various properties of different objects exhibit challenges to both RGB and depth modalities. The evaluation closely follows the protocols adopted in comparison methods \cite{xiang2017posecnn, deng2019poserbpf, li2018deepim, wang2019densefusion} and reports the AUC (Area Under Curve) results on the keyframes in 12 video test sequences evaluated by the metrics of $ADD = \frac{1}{m}\sum_{x \in M} ||Rx+T-(\hat{R}x+\hat{T})||$ which performs exact model matching, and $ADD$-$S = \frac{1}{m}\sum_{x_1 \in M} \min_{x_2 \in M}||Rx_1+T-(\hat{R}x_2+\hat{T})||$ \cite{xiang2017posecnn} designed for evaluating symmetric objects, of which the matching between points can be ambiguous for some views.


Although this dataset contains pose annotated training and validation data collected in real world, the proposed approach does not use any of them but is trained only on synthetic data generated by aforementioned pipeline.

\noindent \textbf{YCBInEOAT Dataset} There have been several public benchmarks \cite{xiang2017posecnn,choi2013rgb} where videos are collected by placing the objects statically on a table-top while a camera is moved around to imitate a 6D object pose tracking scenario. This can be limiting for evaluating 6D pose tracking since in a static environment, the entire image can be leveraged to solve for the trajectory of the camera, from which object pose can be inferred \cite{mur2015orb, hodan2017tless}. Additionally, in such scenarios, extreme object rotations, such as out-of-image-plane flipping are less likely to happen than a free moving object in front of the camera. Thus, exclusive evaluation on such datasets cannot entirely reflect the attributes of a 6D object pose tracking approach. Other datasets \cite{issac2016depth, krull20146} collected video sequences where objects are manipulated by a human hand. Nevertheless, human arm and hand motions can greatly vary from those of robots.

\begin{figure}[t]
  \centering
  \includegraphics[width=0.45\textwidth]{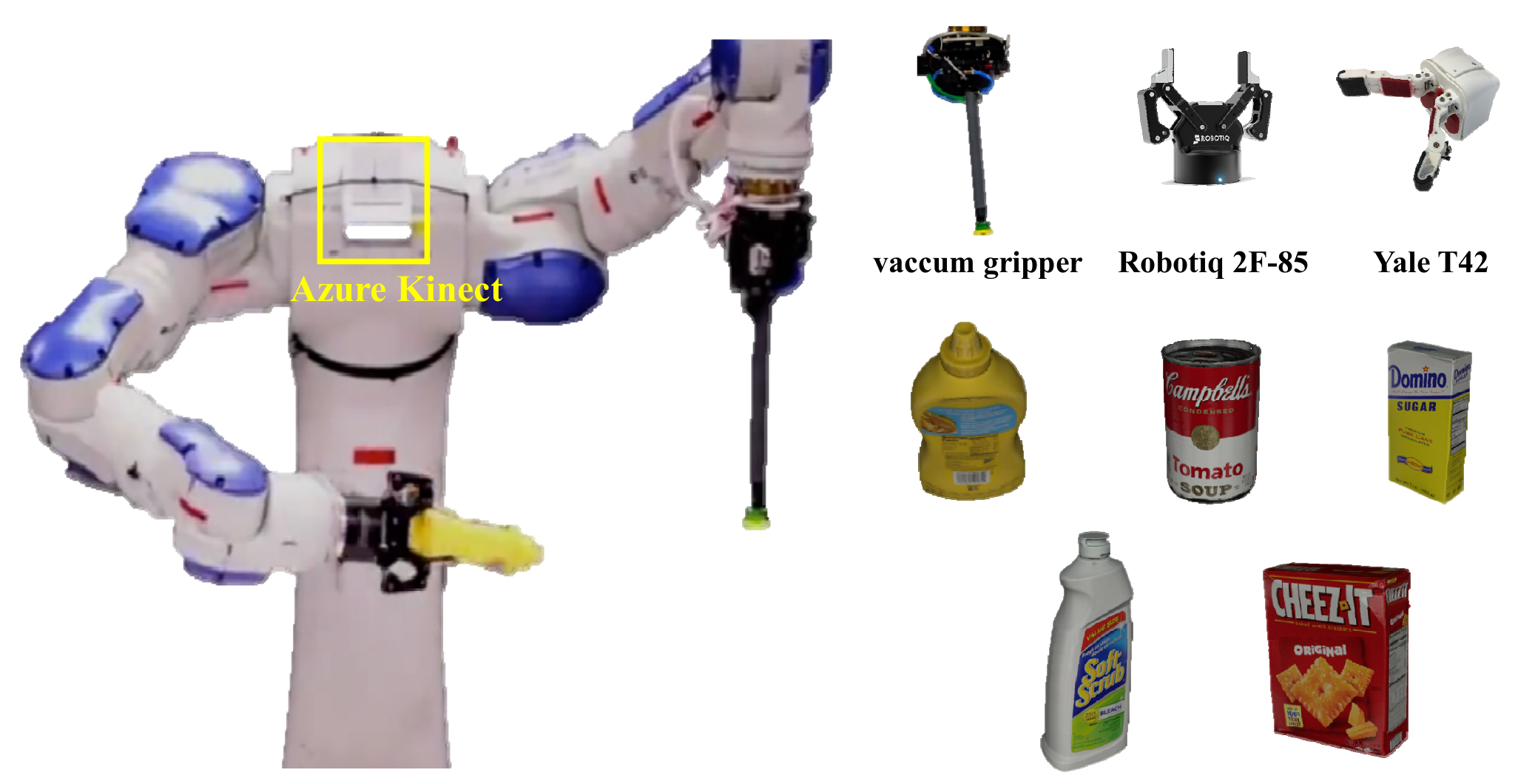}
  \vspace{-0.15in}
  \caption{\textbf{Left:} The dataset collection setup, where manipulation task is performed on a {\it Yaskawa Motoman SDA10f}. \textbf{Right:} Different end-effector modalities and YCB objects that have been used for manipulation.\vspace{-0.1in}}
  \label{fig:setup}
  \vspace{-0.1in}
\end{figure}

Therefore, in this work a novel dataset, referred to as "YCBInEOAT Dataset", is developed in the context of robotic manipulation, where various robot end-effectors are included: a vacuum gripper, a Robotiq 2F-85 gripper, and a Yale T42 Hand \cite{odhner2013open}. The manipulation sequences consider 5 YCB objects, given that they are widely accessible. The data collection setup and objects are shown in Fig. \ref{fig:setup}. Each video sequence is collected from a real manipulation performed with a dual-arm {\it Yaskawa Motoman SDA10f}. In general, there are 3 types of manipulation tasks performed: (1) single arm pick-and-place, (2) within-hand manipulation, and (3) pick to hand-off between arms to placement. RGB-D images are captured by an {\it Azure Kinect} sensor mounted statically on the robot with a frequency of 20 to 30 Hz. Similar to the YCB-Video, ADD and ADD-S metrics are adopted for evaluation. Ground-truth 6D object poses in camera's frame have been accurately annotated \emph{manually} for each frame of the video. The extrinsic parameters of the camera in the frame of the robot has been obtained by a calibration procedure. This dataset is available for future benchmarking.\footnote{https://github.com/wenbowen123/iros20-6d-pose-tracking}

\begin{table*}[]
\centering
\resizebox{0.95\textwidth}{!}{%
\begin{tabular}{|c|cc|cc|cc|cc|cc|cc|aa||cc|aa|aa|}
\hline
                          & \multicolumn{2}{c|}{DOPE \cite{tremblay2018deep}}      & \multicolumn{2}{c|}{DenseFusion \cite{wang2019densefusion}} & \multicolumn{2}{c|}{PoseCNN+ICP+DeepIM \cite{li2018deepim}} & \multicolumn{2}{c|}{DeepIM tracking \cite{li2018deepim}} & \multicolumn{2}{c|}{RGF \cite{issac2016depth}}         & \multicolumn{2}{c|}{Wüthrich's \cite{Wthrich2013ProbabilisticOT}}  & \multicolumn{2}{a||}{$se(3)$-TrackNet}        & \multicolumn{2}{c|}{PoseRBPF \cite{deng2019poserbpf}} & \multicolumn{2}{a|}{$se(3)$-TrackNet}     & \multicolumn{2}{a|}{$se(3)$-TrackNet}     \\ \hline
Modality                  & \multicolumn{2}{c|}{RGB}       & \multicolumn{2}{c|}{RGBD}        & \multicolumn{2}{c|}{RGBD}               & \multicolumn{2}{c|}{RGBD}            & \multicolumn{2}{c|}{Depth}       & \multicolumn{2}{c|}{Depth}       & \multicolumn{2}{a||}{RGBD}        & \multicolumn{2}{c|}{RGBD}     & \multicolumn{2}{a|}{RGBD}     & \multicolumn{2}{a|}{RGBD}     \\
Type                      & \multicolumn{2}{c|}{detection} & \multicolumn{2}{c|}{detection}   & \multicolumn{2}{c|}{detection}          & \multicolumn{2}{c|}{tracking}        & \multicolumn{2}{c|}{tracking}    & \multicolumn{2}{c|}{tracking}    & \multicolumn{2}{a||}{tracking}    & \multicolumn{2}{c|}{tracking} & \multicolumn{2}{a|}{tracking} & \multicolumn{2}{a|}{tracking} \\
Initial pose from         & \multicolumn{2}{c|}{-}         & \multicolumn{2}{c|}{-}           & \multicolumn{2}{c|}{-}                  & \multicolumn{2}{c|}{groundtruth}     & \multicolumn{2}{c|}{groundtruth} & \multicolumn{2}{c|}{groundtruth} & \multicolumn{2}{a||}{groundtruth} & \multicolumn{2}{c|}{PoseCNN}  & \multicolumn{2}{a|}{PoseCNN}  & \multicolumn{2}{a|}{PoseCNN}  \\
Re-initialization (Total) & \multicolumn{2}{c|}{No}        & \multicolumn{2}{c|}{No}          & \multicolumn{2}{c|}{No}                 & \multicolumn{2}{c|}{Yes (290)}       & \multicolumn{2}{c|}{No}          & \multicolumn{2}{c|}{No}          & \multicolumn{2}{a||}{No}          & \multicolumn{2}{c|}{Yes (2)}  & \multicolumn{2}{a|}{No}       & \multicolumn{2}{a|}{Yes (2)}  \\ \hline
Train data                & \multicolumn{2}{c|}{Syn}       & \multicolumn{2}{c|}{Real+Syn}    & \multicolumn{2}{c|}{Real+Syn}           & \multicolumn{2}{c|}{Real+Syn}        & \multicolumn{2}{c|}{-}           & \multicolumn{2}{c|}{-}           & \multicolumn{2}{a||}{Syn}         & \multicolumn{2}{c|}{Syn}      & \multicolumn{2}{a|}{Syn}      & \multicolumn{2}{a|}{Syn}      \\ \hline
Objects                   & ADD            & ADD-S         & ADD            & ADD-S           & ADD                & ADD-S              & ADD               & ADD-S            & ADD              & ADD-S         & ADD             & ADD-S          & ADD             & ADD-S          & ADD           & ADD-S         & ADD           & ADD-S         & ADD           & ADD-S         \\ \hline
002\_master\_chef\_can    &                & -             & -              & 96.40           & 78.00              & 96.30              & 89.00             & 93.80            & 46.23            & 90.17         & 55.60           & 90.68          & 93.86           & 96.29          & 90.50         & 95.10         & 93.84         & 95.92         & 93.84         & 95.92         \\
003\_cracker\_box         & 55.90          & 69.80         & -              & 95.50           & 91.40              & 95.30              & 88.50             & 93.00            & 56.95            & 72.26         & 96.38           & 97.19          & 96.52           & 97.20          & 88.20         & 93.00         & 96.42         & 97.12         & 96.42         & 97.12         \\
004\_sugar\_box           & 75.70          & 87.10         & -              & 97.5            & 97.60              & 98.20              & 94.30             & 96.30            & 50.38            & 72.65         & 97.14           & 97.94          & 97.58           & 98.14          & 92.90         & 95.50         & 97.56         & 98.13         & 97.56         & 98.13         \\
005\_tomato\_soup\_can    & 76.10          & 85.10         & -              & 94.60           & 90.30              & 94.80              & 89.10             & 93.20            & 72.44            & 91.60         & 64.74           & 89.55          & 94.96           & 97.17          & 90.00         & 93.80         & 94.81         & 97.10         & 94.81         & 97.10         \\
006\_mustard\_bottle      & 81.90          & 90.90         & -              & 97.20           & 97.10              & 98.00              & 92.00             & 95.10            & 87.71            & 98.19         & 97.12           & 97.95          & 95.76           & 97.37          & 91.90         & 96.30         & 95.73         & 97.36         & 95.73         & 97.36         \\
007\_tuna\_fish\_can      & -              & -             & -              & 96.60           & 92.20              & 98.00              & 92.00             & 96.40            & 28.67            & 52.93         & 69.14           & 93.32          & 86.46           & 91.09          & 91.10         & 95.30         & 86.46         & 91.08         & 86.46         & 91.08         \\
008\_pudding\_box         & -              & -             & -              & 96.50           & 83.50              & 90.60              & 80.10             & 88.30            & 12.69            & 17.98         & 96.85           & 97.89          & 97.93           & 98.39          & 85.80         & 92.00         & 97.90         & 98.37         & 97.90         & 98.37         \\
009\_gelatin\_box         & -              & -             & -              & 98.10           & 98.00              & 98.50              & 92.00             & 94.40            & 49.10            & 70.72         & 97.46           & 98.37          & 97.81           & 98.42          & 96.30         & 97.50         & 97.74         & 98.46         & 97.74         & 98.46         \\
010\_potted\_meat\_can    & 39.40          & 52.40         & -              & 91.30           & 82.20              & 90.30              & 78.00             & 88.90            & 44.09            & 45.57         & 83.71           & 86.69          & 77.81           & 84.16          & 68.70         & 77.90         & 36.45         & 60.28         & 74.51         & 82.38         \\
011\_banana               & -              & -             & -              & 96.60           & 94.90              & 97.60              & 81.00             & 90.50            & 93.33            & 97.74         & 86.27           & 96.07          & 94.90           & 97.18          & 74.20         & 86.90         & 40.04         & 78.81         & 84.62         & 95.15         \\
019\_pitcher\_base        & -              & -             & -              & 97.10           & 97.40              & 97.90              & 90.40             & 94.70            & 97.93            & 98.18         & 97.30           & 97.74          & 96.75           & 97.45          & 86.80         & 94.20         & 96.71         & 97.43         & 96.71         & 97.43         \\
021\_bleach\_cleanser     & -              & -             & -              & 95.80           & 91.60              & 96.90              & 81.70             & 90.50            & 95.87            & 97.28         & 95.23           & 97.16          & 95.94           & 97.25          & 86.00         & 93.00         & 95.89         & 97.23         & 95.89         & 97.23         \\
024\_bowl                 & -              & -             & -              & 88.20           & 8.10               & 87.00              & 38.80             & 90.60            & 24.25            & 82.40         & 30.37           & 97.15          & 80.91           & 94.46          & 25.50         & 94.20         & 39.12         & 95.56         & 39.12         & 95.56         \\
025\_mug                  & -              & -             & -              & 97.10           & 94.20              & 97.60              & 83.20             & 92.00            & 59.99            & 71.18         & 83.15           & 93.35          & 91.53           & 96.88          & 90.90         & 97.10         & 91.56         & 96.88         & 91.56         & 96.88         \\
035\_power\_drill         & -              & -             & -              & 96.00           & 97.20              & 97.90              & 85.40             & 92.30            & 97.94            & 98.35         & 97.09           & 97.82          & 96.42           & 97.40          & 93.90         & 96.10         & 96.38         & 97.38         & 96.38         & 97.38         \\
036\_wood\_block          & -              & -             & -              & 89.70           & 81.10              & 91.50              & 44.30             & 75.40            & 45.68            & 62.51         & 95.48           & 96.87          & 95.16           & 96.70          & 20.10         & 89.10         & 33.91         & 95.92         & 33.91         & 95.92         \\
037\_scissors             & -              & -             & -              & 95.20           & 92.70              & 96.00              & 70.30             & 84.50            & 20.94            & 38.60         & 4.17            & 16.20          & 95.68           & 97.55          & 76.10         & 85.60         & 95.67         & 97.54         & 95.67         & 97.54         \\
040\_large\_marker        & -              & -             & -              & 97.50           & 88.90              & 98.20              & 80.40             & 91.20            & 12.17            & 18.90         & 35.58           & 53.02          & 92.15           & 95.99          & 92.00         & 97.10         & 89.01         & 94.23         & 89.01         & 94.23         \\
051\_large\_clamp         & -              & -             & -              & 72.90           & 54.20              & 77.90              & 73.90             & 84.10            & 62.84            & 80.12         & 61.25           & 72.35          & 94.71           & 96.93          & 48.50         & 94.80         & 71.60         & 96.88         & 71.60         & 96.88         \\
052\_extra\_large\_clamp  & -              & -             & -              & 69.80           & 36.50              & 77.80              & 49.30             & 90.30            & 67.48            & 69.65         & 93.73           & 96.58          & 91.74           & 95.76          & 40.30         & 90.10         & 64.58         & 95.80         & 64.58         & 95.80         \\
061\_foam\_brick          & -              & -             & -              & 92.50           & 48.20              & 97.60              & 91.60             & 95.50            & 69.99            & 86.55         & 96.76           & 98.11          & 93.65           & 96.71          & 81.10         & 95.70         & 40.66         & 94.67         & 40.66         & 94.67         \\ \hline
ALL                       & -              & -             & -              & 93.10           & 80.70              & 94.00              & 79.30             & 91.00            & 59.18            & 74.29         & 78.01           & 90.21          & 93.05           & 95.71          & 80.80         & 93.30         & 84.46         & 93.87         & 87.81         & 95.52         \\ \hline
Speed (fps)               & \multicolumn{2}{c|}{4.31}      & \multicolumn{2}{c|}{16.67}       & \multicolumn{2}{c|}{0.09}               & \multicolumn{2}{c|}{12.00}           & \multicolumn{2}{c|}{11.76}       & \multicolumn{2}{c|}{12.93}       & \multicolumn{2}{a||}{90.90}       & \multicolumn{2}{c|}{5.00}     & \multicolumn{2}{a|}{90.90}    & \multicolumn{2}{a|}{90.90}    \\ \hline
\multicolumn{21}{p{27cm}}{\ \vspace{-0.1in} \newline \large Table I: Comparing the performance of $se(3)$-TrackNet (Gray) with state-of-the-art techniques on the {\it YCB Video}. The approach significantly outperforms the competing approaches over the ADD metric, which considers semantic information during pose evaluation. It also achieves the highest success rate over the ADD-S metric both in cases of initialization with the ground-truth pose and when initialized with the output of PoseCNN \cite{xiang2017posecnn} (rightmost two columns).}
\end{tabular}%
}
\label{tab:ycb_res}
\end{table*}

\begin{figure*}[t]
  \centering
  \includegraphics[width=0.95\textwidth]{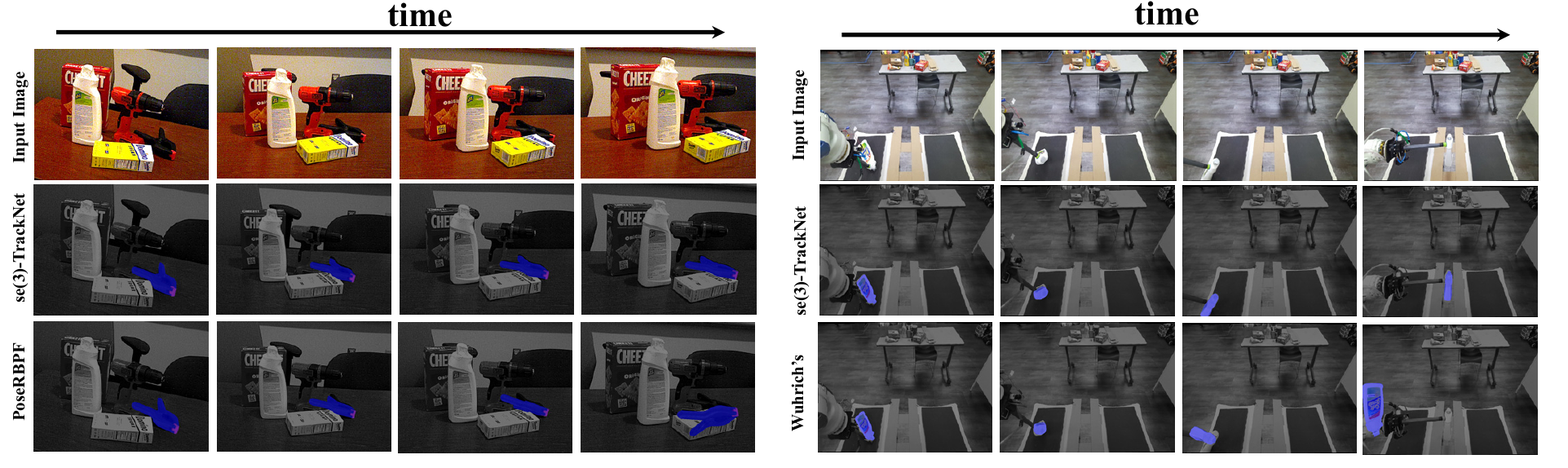}
  \vspace{-0.2in}
  \caption{\textbf{Left:} Qualitative results for tracking the  "large-clamp" object in the YCB-Video dataset. \textbf{Right:} Tracking results for "bleach-cleanser" being manipulated by a vacuum gripper in the {\it YCBInEOAT} dataset.\vspace{-0.1in}}
  \vspace{-0.in}
  \label{fig:qual_tomato}
\end{figure*}

\subsection{Results on YCB-Video}

Table I and Fig. \ref{fig:acc_speed} present the evaluation over the {\it YCB-Video} dataset. The proposed approach is compared with other state-of-art 6D object pose detection approaches   \cite{xiang2017posecnn, tremblay2018deep, li2018deepim, wang2019densefusion} and 6D pose tracking approaches \cite{deng2019poserbpf, li2018deepim, issac2016depth, Wthrich2013ProbabilisticOT}, where publicly available source code\footnote{https://github.com/bayesian-object-tracking/dbot} is used to evaluate \cite{issac2016depth,Wthrich2013ProbabilisticOT}, while other results are adopted from the respective publications. All the compared tracking methods except PoseRBPF are using ground-truth pose for initialization. PoseRBPF \cite{deng2019poserbpf} is the only one that is initialized using predicted poses from PoseCNN \cite{xiang2017posecnn}. For fairness, two additional experiments using the same initial pose as PoseRBPF\footnote{We thank the authors for providing the initial pose they used in the original paper \cite{deng2019poserbpf}} are performed and presented in the rightmost two columns of Table I, one is without any re-initialization, and the other allows re-initialization by PoseCNN twice (same as in PoseRBPF) after heavy occlusions. The prior work \cite{li2018deepim} was originally proposed to refine the pose output from any 6D pose estimation detection method, but also extends to RGB-based tracking. It has to be re-initialized by PoseCNN when the last 10 frames have an average rotation greater than 10 degrees or an average translation greater than 1 cm, which happens every 340 frames on average as reported \cite{li2018deepim}. The initial pose is from ground-truth. 

In practice, re-initialization can be quite expensive in robotics applications given the slower running speed of 6D pose detection approaches, which can interrupt and adversely affect other components of the system, such as planning and control. It can also introduce new source of error. In contrast, the proposed network performs long-term, accurate tracking with less frequent or even no re-initialization while trained using only on synthetic data. Additionally, the proposed method generalizes to different lighting conditions and a variety of objects with different properties, such as {\it scissors}, and {\it clamps}, which are challenging to alternatives. Another important aspect is that the proposed approach is able to achieve {\bf 93.05\%} on ADD metric, outperforming all comparison methods by a large margin. This can be attributed to its implicitly learnt residual estimator, which not only captures the discrepancy of geometry but also semantic textures by considering both the RGB and Depth modalities. 

\subsection{Results on YCBInEOAT-Dataset}
\begin{table}[]
\centering
\resizebox{0.48\textwidth}{!}{%
\begin{tabular}{ccccccc}
\hline 
Objects & \multicolumn{2}{c}{RGF \cite{issac2016depth}} & \multicolumn{2}{c}{Wüthrich's \cite{Wthrich2013ProbabilisticOT}} & \multicolumn{2}{c}{$se(3)$-TrackNet} \\
& ADD       & ADD-S       & ADD         & ADD-S         & ADD         & ADD-S      \\
\hline                    
003\_cracker\_box     & 34.78          &  55.44           & 79.00            &   88.13            & 90.76       & 94.06      \\
021\_bleach\_cleanser &  29.40         &  45.03           &  61.47           &  68.96             & 89.58       & 94.44      \\
004\_sugar\_box       &  15.82         &  16.87           &  86.78           &   92.75            & 92.43       & 94.80      \\
005\_tomato\_soup\_can &  15.13         &  26.44           &  63.71           &   93.17            & 93.40       & 96.95      \\
006\_mustard\_bottle  &   56.49        &  60.17           &  91.31           &  95.31             & 97.00       & 97.92      \\
\hline 
ALL                 &  29.98         & 39.90            &  78.28           &   89.18            & 92.66       & 95.53\\
\hline
\multicolumn{7}{p{9cm}}{\vspace{0.01in}Table II: Results evaluated on YCBInEOAT-dataset by AUC (Area Under Curve) for ADD and ADD-S.\vspace{-1in}}
\end{tabular}%
}
\label{tab:mydata_res}
\vspace{-0.1in}
\end{table}

Table II shows the quantitative results evaluated by the area under the curve for ADD and ADD-S on the developed {\it YCBInEOAT} dataset. On this benchmark, the tracking approaches with publicly available source code could be directly evaluated \cite{Wthrich2013ProbabilisticOT,issac2016depth}. Pose is initialized with ground-truth in the first frame and no re-initialization is allowed. Forward kinematics are not leveraged in order to solely evaluate tracking quality. Example qualitative results are demonstrated by Fig. \ref{fig:qual_tomato} where a vacuum gripper is performing a pick-and-place manipulation task. Abrupt motions, extreme rotations and slippage within the end-effector are introduced, which are challenging for 6D object pose tracking. Nevertheless, the proposed approach is sufficiently robust to provide long-term reliable pose estimation until the end of the manipulation.

\label{sec:ablation}
\subsection{Ablation Study}

\begin{wrapfigure}{r}{0.5\linewidth}
\vspace{-0.1in}

\resizebox{0.245\textwidth}{!}{%
\begin{tabular}{c|c|c}
\hline
Criteria & ADD   & ADD-S \\ \hline
Proposed             & 94.71 & 96.93 \\
No physics       & 91.88 & 95.76 \\
No depth         & 75.65 & 87.22 \\
Shared encoder   & 0.28  & 0.28  \\
Quaternion       & 93.58 & 96.39 \\
Shape-Match Loss & 1.93  & 5.48  \\ \hline
\multicolumn{3}{p{6cm}}{Ablation study on critical components of our framework.}
\end{tabular}%
}
\label{tab:ablation}
\vspace{-0.17in}
\end{wrapfigure}
An ablation study investigates the importance of different modules of the proposed approach. It is performed for the {\it large clamp} object from the {\it YCB-Video dataset} and is presented in the accompanying table. The initial pose is given by ground-truth and no re-initialization is allowed. \textbf{No physics} implies that domain randomization is employed during synthetic data generation without any physics simulation. For \textbf{No depth}, the depth modality is removed in both training and inference stage to study its importance. 
\textbf{Shared encoder} means the two feature encoders for $I^t$ and $I^{t-1}$ share the same architecture and weights. This corresponds to the one used for $I^{t-1}$ in the original setup. \textbf{Quaternion} implements the rotation via a quaternion representation $q=(x,y,z,w)$, where $w=\sqrt{1-x^2-y^2-z^2}$ is forced to be non-negative so as to avoid the ambiguity of $q$ and $-q$. The network is trained by $L_2$ loss over the representation. \textbf{Shape-Match Loss} is a popular loss function in 6D pose estimation task that does not require the specification of symmetries \cite{xiang2017posecnn}. It loses track, however, of the object very early in the current setting.

\section{CONCLUSION}
This work presents a framework for efficient and robust long-term 6D object pose tracking. A novel neural network architecture $se(3)$-TrackNet is proposed that allows training on synthetic datasets that transfers robustly to real world data. A combination of design choices for the network and the Lie Algebra representation for learning residual poses during pose tracking result in highly desirable performance validated by extensive experiments. The pose tracking process operates at approximately 90.90 fps, which is significantly higher than alternatives. An additional dataset is proposed to address the lack of an object tracking benchmark in the robotics manipulation context. $se(3)$-TrackNet is shown to be robust under large occlusions and sudden re-orientations introduced in the dataset, which challenge competing approaches. Despite these desirable properties for the proposed network, a limitation is that an object CAD model is required. The future objective is to achieve similar performance for category-level 6D pose tracking.


\bibliographystyle{IEEEtran}
\bibliography{ref.bib}

\end{document}